\documentclass{article}

\usepackage{algorithmic}  
\usepackage[ruled,linesnumbered]{algorithm2e}
\usepackage{CJK}
\usepackage{stfloats}
\usepackage{graphicx} %插入图片的宏包
\usepackage{float} %设置图片浮动位置的宏包
\usepackage{subfigure} %插入多图时用子图显示的宏包
\usepackage{cite}
\usepackage{amsmath}
\usepackage{authblk}

\title{A multi-perspective combined recall and rank framework for Chinese procedure terminology normalization}
\author{Ming Liang}
\author{Kui Xue}
\author{Tong Ruan\thanks{Corresponding author: ruanton@mail.ecust.edu.cn}}
\affil{East China University of Science and Technology}

\begin{document}

\maketitle

\begin{abstract}
Medical terminology normalization aims to map the clinical mention to terminologies come from a knowledge base, which plays an important role in analyzing Electronic Health Record(EHR) and many downstream tasks. In this paper, we focus on Chinese procedure terminology normalization. The expression of terminologies are various and one medical mention may be linked to multiple terminologies. Previous study explores some methods such as multi-class classification or learning to rank(LTR) to sort the terminologies by literature and semantic information. However, these information is inadequate to find the right terminologies, particularly in multi-implication cases. In this work, we propose a combined recall and rank framework to solve the above problems. This framework is composed of a multi-task candidate generator(MTCG), a keywords attentive ranker(KAR) and a fusion block(FB). MTCG is utilized to predict the mention implication number and recall candidates with semantic similarity. KAR is based on Bert with a keywords attentive mechanism which focuses on keywords such as procedure sites and procedure types. FB merges the similarity come from MTCG and KAR to sort the terminologies from different perspectives. Detailed experimental analysis shows our proposed framework has a remarkable improvement on both performance and efficiency.\\
\end{abstract}
\newpage

\section{Introduction}
Mining and exploring structured data from Electronic Health Record(EHR) plays a key role in many applications such as clinical research \cite{gonzalez2016recent,fleuren2015application} and decision support systems \cite{rumshisky2016predicting,liu2019towards}. However, the non-standard and various expression of medical terminologies hinders the further utilization of EHR. For example, "\begin{CJK*}{UTF8}{gbsn}甲状腺病损切除术\end{CJK*}" could be described as "\begin{CJK*}{UTF8}{gbsn}右侧甲状腺肿瘤切除术\end{CJK*}", "\begin{CJK*}{UTF8}{gbsn}甲状腺肿物切除术\end{CJK*}" and "\begin{CJK*}{UTF8}{gbsn}腔镜双侧甲状腺包块切除术\end{CJK*}". If the analysis is directly based on such data, a lot of valid information will be ignored and wasted. Therefore description of medical terminologies should be normalized before analyzing EHR. Medical terminology normalization aims to link the description of medical terminology in EHR to standard entities in a knowledge base. Among all the terminologies, the descriptions of procedure are most diverse and requires professional skills. Clinicians often add some extra description such as procedure instruments and concrete procedure steps, most of which are irrelevant for normalization but useful in describing the process of procedure. In this paper, we focus on Chinese procedure terminology normalization. The procedure terminologies come from International Classification of Disease version 9(ICD-9). 

\begin{figure*}[!t]
    \centerline{\includegraphics{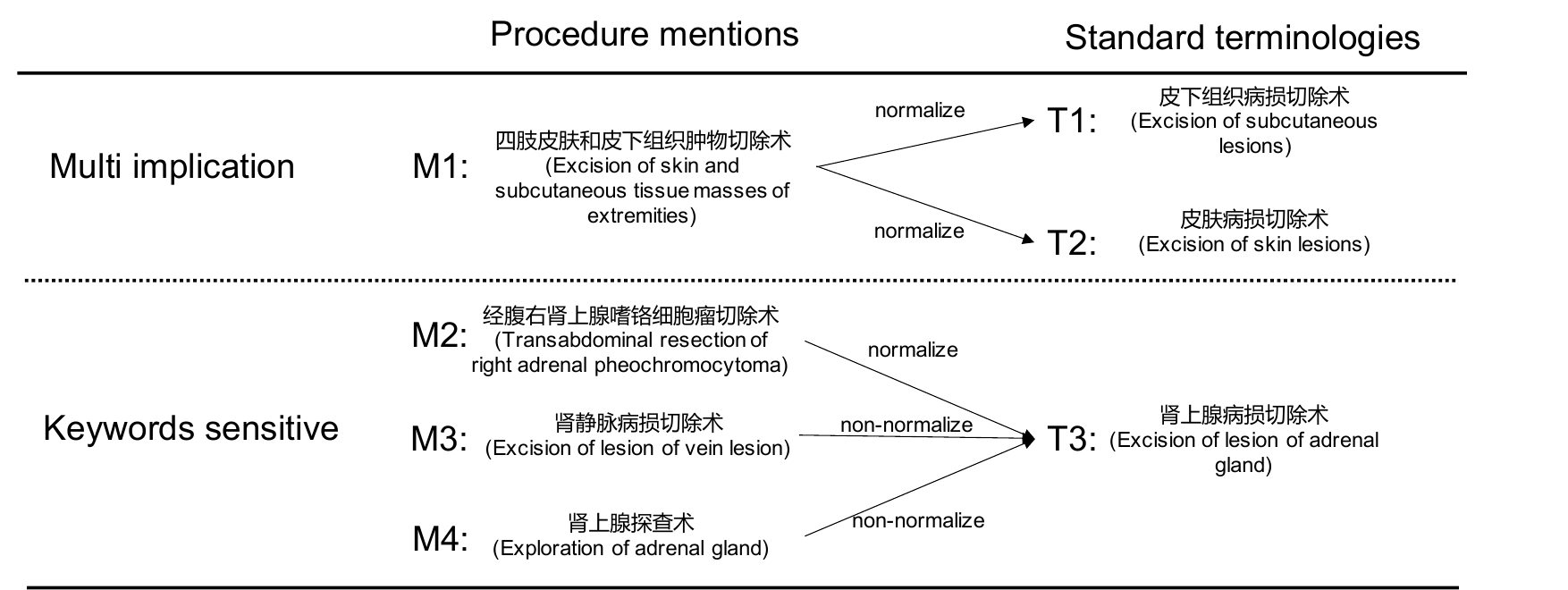}}
    \caption{Examples of multi implication and keywords sensitive}
    \label{fig:figure1}
\end{figure*}

We notate the statement of procedure written by doctors in EHR as procedure mention and its corresponding entities in knowledge base as terminologies. The characteristics and challenges of Chinese procedure terminology normalization are: \textbf{1) Multi implication.} As proposed in \cite{yan2020knowledge}, multi implication means a procedure mention may be normalized to multiple terminologies. An example is shown in the first part of Figure 1. M1 maps to two terminologies because two procedure sites "\begin{CJK*}{UTF8}{gbsn}皮肤\end{CJK*}(skin)" and "\begin{CJK*}{UTF8}{gbsn}皮下组织\end{CJK*}(Subcutaneous tissue)" are presented in one procedure mention. \textbf{2)Short text.} Unlike entity linking or entity normalization, no context information is available in this task. Both procedure mentions and terminologies are short text, with an average length of 12 and 9 characters in Chinese respectively. \textbf{3)Keywords sensitive.} Keywords such as procedure site and procedure type matter in terminology normalization. As shown in the second part of Figure 1, M2 matches with T3 because "\begin{CJK*}{UTF8}{gbsn}嗜铬细胞瘤\end{CJK*}(pheochromocytoma)" is a sub-concept of "\begin{CJK*}{UTF8}{gbsn}病损\end{CJK*}(lesion)" and "\begin{CJK*}{UTF8}{gbsn}经腹\end{CJK*}(Transabdominal)" is just an extra description which has no affect on the result. M3 and M4 are not normalized to T3 because they share different procedure site and procedure type respectively. \textbf{4)High efficiency.} Knowledge base usually has a large amount of terminologies, the designed algorithm should generate normalized terminologies quickly and efficiently.

Previous studies on medical terminology normalization could be divided into two categories: "direct rank" and "rank follows recall". Direct rank based methods such as string match, dictionary look up \cite{leal2015ulisboa,d2015sieve,lee2016audis}, multi-class classification \cite{miftahutdinov2019deep,luo2019hybrid,niu2019multi} and point-wise learning to rank \cite{deng2019ensemble,luo2018multi} directly select the mapping terminologies from the whole knowledge base. Among them, methods such as string match and dictionary look up cannot process synonyms that are literally different but semantically the same; The output space of multi-class classification is the same as the number of terminologies in the knowledge base. Multi-class classification approaches may not perform well on the terminologies which have not been appeared in the training data. Point-wise learning to rank regards terminology normalization as a binary classification problem. The input of the model is a medical mention with a terminology, the output is the similarity of the two texts. With the size of knowledge base increases, point-wise learning to rank methods faces serious efficiency problem. Rank follows recall approaches \cite{mondal2020medical,liang2020lab,ji2020bert,li2017cnn,zhang2018effective} applies a two-step framework which first generate candidate terminologies by heuristic rules or statistic methods then rank the candidates in a point-wise way. The recall step only decreases the scale of the candidates without providing other information for sorting the candidates. The similarity calculated in recall step is often wasted. Thus the output terminologies could only be sorted during the rank step. We find that the recall and rank step should sort the input terminologies from different perspectives, then combine their results to complement each other. Besides, all of the methods mentioned above could not solve multi-implication problem except\cite{yan2020knowledge}, which utilize a sequence generation model to generate all possible terminologies with the cost of losing efficiency.

In this paper, we propose a novel "combined recall and rank" framework with a multi-task candidate generator(MTCG), a keywords attentive ranker(KAR) and a fusion block(FB). We first recall literature and semantic similar terminologies by a pairwise recall model. Then rank these candidates according to context and keywords information. Finally the similarities come from recall and rank step are merged to generate the normalized results. In contrast with previous pairwise model, we propose an effective online negative sampling strategy which greatly improves recall rate. And compared with other rank model, keywords such as procedure site and procedure type are fully utilized for better performance. To handle multi-implication problem, we define it as a classification task and train it with the recall model at the same time.

Overall, the contributions of this paper are as follows:

\begin{itemize}
    \item We propose a novel "combined recall and rank" framework for Chinese procedure terminology normalization. A fusion block is designed to merge the results from different perspective of recall and rank model.
    \item We design a multi-task candidate generator aims to predict both the implication number and recall the candidate terminologies. An effective online negative sampling strategy is proposed to find informative negative samples for training. To the best of our knowledge, negative sample strategy is first considered in medical terminology normalization. 
    \item We introduce a keywords attentive ranker, which focuses on the procedure site and procedure type of mentions and terminologies. The attention on keywords provides another perspective for sorting the terminologies
\end{itemize}

%\enlargethispage{12pt}
\section{Related work}
There are two areas related to our work: medical terminology normalization and text matching. Text matching aims to infer the relation of two given texts. In this task, text matching methods could be used to prediction the similarity of procedure mentions and terminologies.

\subsection{Medical terminology normalization}
Medical terminology normalization aims to find standard terminologies for a medical mention, usually the terminologies come from a knowledge base or a terminology dictionary. Most of the proposed methods could divided into two categories: "direct rank" and "rank follows recall". Direct rank directly finds the normalized terminologies from the whole knowledge base. Early methods \cite{leal2015ulisboa} and \cite{d2015sieve} match medical mention to terminologies by heuristic rules.With the appearance of machine learning, \cite{miftahutdinov2019deep, niu2019multi,belousov2017using,limsopatham2016normalising} regard medical terminology normalization as a multi-class classification problem where the number of categories is the same as the number of terminologies. \cite{niu2019multi} designs a multi-task framework which normalize both mentions and abbreviations at the same time. \cite{limsopatham2016normalising} utilize recurrent neural network(RNN) to capture the semantic meaning of medical mentions and terminologies. Point-wise learning to rank is also used to normalize terminologies. \cite{luo2018multi} propose a multi-view convolutional neural network(CNN) and a multi-task framework to normalize both procedure and disease mentions. When the terminology knowledge base is large, \cite{liang2020lab,ji2020bert,li2017cnn,xu2020generate,yan2020knowledge,zhang2018effective} propose a recall model to generate possible terminologies then followed by a rank model to sort. \cite{ji2020bert,li2017cnn} first generate candidates by bm25, then rank the terminologies by CNN and Bert respectively. \cite{liang2020lab} applies active learning on lab indicator normalization when training the binary classification rank model, which aims to reduce the cost of annotation. \cite{xu2020generate} design a list-wise model with semantic type regulation to sort the terminologies. \cite{zhang2018effective} recalls candidates by clustering then sort them based on ngram features. \cite{yan2020knowledge} design a sequence generate and rank framework which could solve multi-implication problem, but they ignore the efficiency problem.

\subsection{Text matching}
Text matching aims to predict the relevance of two given texts. In this task, text matching is applied to judge the similarity of medical mentions and terminologies. The proposed methods in text matching could be divided into two categories\cite{khattab2020colbert}: representation based and interaction based. Representation based methods encode text into vectors and calculate vector similarity as their similarity. \cite{mueller2016siamese}, \cite{zhou2016multi}, and \cite{reimers2019sentence} utilize different encoder structure such as LSTM, CNN and Bert respectively. Representation based methods are efficient but there is no interaction between the two input sentences. With the appearance of attention mechanism, a lot of works have been focused on interaction based methods\cite{rajagopalan2016extending,wan2016match,xiong2017end}. \cite{chen2016enhanced} uses bidirectional LSTM to encode the sentences and applies cross attention between two sentences. \cite{wang2017bilateral} utilizes an advanced bilateral multi-perspective matching for natural language sentences. Bert \cite{devlin2018bert} is also used in this area and performs well on many tasks. \cite{qiao2019understanding} studies the behaviors of Bert on ranking task. \cite{khattab2020colbert} explore a late interaction formula based on Bert which contains both representation and interaction. \cite{yang2019simple} summaries the function of the components in interactive based model and proposes three key features for alignment. \cite{peng2020enhanced} proposes an Enhanced-RCNN for learning sentence similarity which is far less complex compared with Bert.

\section{Materials and methods}
The whole architecture of our framework is shown in Figure 2. During the training process, we first train MTCG. It encodes procedure mentions and terminologies into high dimensional vectors, their similarities are measured by the euclidean distance. MTCG also predicts mention implication number with captured context information to solve multi implication problems. Candidates of original training set is collected by MTCG for training KAR. During inference time, FB merges the similarities come from different perspective of MTCG and KAR to output the final results.

\begin{figure*}[!t]
    \centerline{\includegraphics{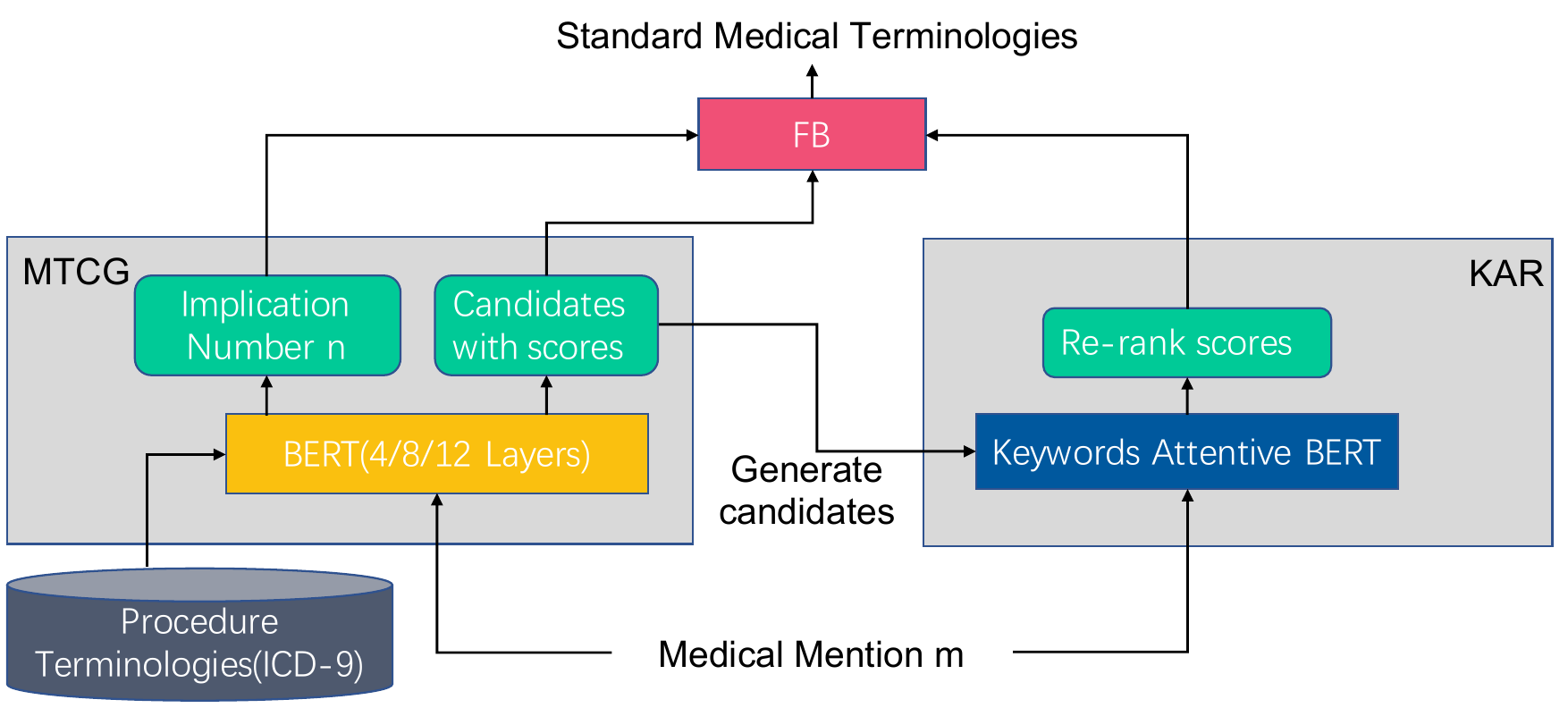}}
    \caption{Combined recall and rank framework. MTCG stands for multi task candidate generator, KAR stands for Keywords attentive ranker, FB stands for fusion block. The input of KAR is generated by MTCG during training and inference time. FB merges the result of MTCG and KAR to output the final terminologies}
    \label{fig:2}
\end{figure*}
\subsection{Multi-task candidate generator}
\subsubsection{Multi-task model}
MTCG handles both a recall task and a mention implication number prediction task. The main idea to recall candidates is deep metric learning, which embeds texts into vector space where the vectors of similar texts are close and vice versa. Compared with other statistic methods, deep metric learning encodes more semantic information into text embedding to search candidates instead of utilizing only literature similarity. Here we regard procedure mention and its corresponding terminologies as similar text pairs. Mention implication number prediction is a regression problem. However, the statistics of mention implication number shows the data follows a long tail distribution in Table \ref{table1}. Therefore we redefine mention implication number prediction as a classification problem and set the total category number as 3. The label of some cases where its implication number is bigger than 3 is all set to 3, which will be adjusted in fusion block during inference time.

The base model used in MTCG is Bert, a multi-layer transformer architecture. The input of each transformer layer is a matrix $E \in (l, d)$, which is the input embedding matrix or the output of last transformer block. And $l$  is the sequence length, $d$  is the input dimension. Then multi-head self-attention is applied for encoding hidden context information, which is composed of multiple self-attention block. The output of multi-head attention is computed by:
$$
MultiHead(E)= [head^1, head^2, ...,head^n]W_o\space\space\space\space\space\space\space\space\space\space\space\space\space\space(2)
$$
$$
head^i = softmax(\frac{Q^iK^{iT}}{\sqrt{d_{head}}})  V^i\space\space\space\space\space\space\space\space\space\space\space\space\space\space(3)
$$
$$
[Q^i,K^i,V^i] = E[W_q^i, W_k^i, W_v^i]\space\space\space\space\space\space\space\space\space\space\space\space\space\space(4)
$$

The three parameter matrix $W_q^i, W_k^i, W_v^i$ of $ith$ head is used to map the input matrix in to hidden space, with the shape of $(d, d_{head})$. And usually $d = n * d_{head}$ where $n$ is the number of heads. That means the $[head^1, head^2, ...,head^n] \in (l, d)$ and $W_o \in (d, d)$.

A feed-forward neural network block is then used to further process the output of multi-head attention. As the shown in formula 5, in which $W_1 \in (d, d_{ff}),W_2 \in (d_{ff}, d), b_1\in(d_{ff}, 1), b_2 \in(d, 1)$. $d_{ff}$ is a hyper parameter. Other components such as residual network and layer normalization are the same as \cite{yan2020knowledge}.
$$
FNN(x) = max(0, xW_1 + b_1)W_2 + b_2\space\space\space\space\space\space\space\space\space\space\space\space\space\space(5)
$$
\begin{figure*}[h]
    \centerline{\includegraphics{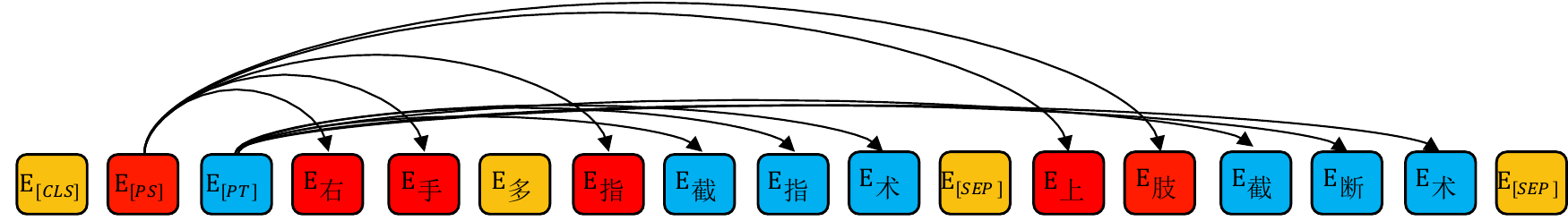}}
    \caption{An example of the input of KAR. Two special token [PS] and [PT] is added. the lines shows the attention of the two tokens. [PS] could only attends to procedure site tokens and [PT] could only attends to procedure type tokens, where the red tokens represent procedure site and blue tokens represent procedure type}
    \label{fig:3}
\end{figure*}

In MTCG, the output of Bert $H \in (l,d)$ is used to represent procedure mentions and terminologies. As shown in Figure 4, During the training process, the input of the model is a procedure mention $M$, a positive terminology $P$ and a negative terminology $N$. For each procedure mention, its standard terminologies are regarded as positive samples and the remained terminologies are all possible negative samples. After encoding by Bert, the sequence hidden output is notated as $H_M, H_P, H_N$. Then an average pooling layer is applied in sequence length dimension to get the vector representation of the input sequence, which is marked as $V_M, V_P, V_N$ respectively. For the recall task, we use triple loss to make the representation of mention closer to its positive samples and farther from its negative samples. And the distance of two vector $D(d_1, d_2)$ is calculated by $||d_1 - d_2||_2$. For the mention implication number prediction task, we take the CLS token of procedure mention $CLS_m$ out, then followed by a FNN and softmax layer. Crossentropy is applied for this classification task. The output of softmax layer is $\hat{y}$ and the $y$ is the implication number label of current mention. The total loss is the sum of triple loss and crossentropy, which is shown in formula 6-8. 
$$
loss = loss_t + loss_c \space\space\space\space\space\space\space\space(6)
$$
$$
loss_t = max(0, D(V_M,V_P) + margin - D(V_M,V_N))\space\space\space\space\space\space\space\space(7)
$$
$$
loss_c = y\log{\hat{y}}\space\space\space\space\space\space\space(8)
$$

\subsubsection{Negative sampling}
As described above, positive samples and negative samples is needed during training candidate generator. The positive samples are fixed and easy to confirm, but there is a large amount of negative samples of a medical mention. For each mention, we choose $k_n$ negative samples for training, where $k_n$ is a hyper parameters. The quality of negative samples directly impacts model performance. The most simple strategy is random sampling. However, random sampling ignores informative samples and does not provide enough information for the model, which normally leads to a poor performance. Thus we design following negative sample strategies:

\begin{algorithm}[h]
\caption{The pseudo-code of online negative sampling}
\label{alg:online negative sampling}
\KwIn{Training set (X, Y), X is medical mentions, Y is its corresponding medical terminologies; Knowledge base KB; An initialized model M; Total training epoch E; Negative sample number of each mention k;}
\KwOut{Model M after training;}
\begin{algorithmic}[1]
\STATE randomly choose negative samples N for X
\FOR{each $e \in [0, E]$}
\STATE training M on (X,Y,N)
\STATE $N \leftarrow \emptyset$ 
\STATE $V_m = M(X)$
\STATE $V_{kb} = M(KB)$
\STATE $S = Euclidean\_distance(V_m * V_{kb})$ 
\STATE $S_{index} = Argsort(S)$ // Ascending order
\FOR{each $i \in |X|$} 
\STATE $n \leftarrow \emptyset$ 
\FOR{each $j \in S_{index}[i]$}
\IF{|n| < k:} 
\IF{$KB[j] \notin Y_j$}
\STATE add KB[j] to n
\ENDIF
\ELSE
\STATE break
\ENDIF
\STATE add n to N
\ENDFOR
\ENDFOR
\ENDFOR

\end{algorithmic}
\end{algorithm}

\begin{itemize}
    \item \textbf{Tf-idf:} We calculate the tf-idf value of procedure mentions and terminologies from knowledge base. And choose the top-$k_n$ negative samples.
    \item \textbf{Tree coding:} The terminologies in ICD9 are roughly categorized by procedures site, which could be conducted by its code. For example, the code of "\begin{CJK*}{UTF8}{gbsn}脑膜切开术\end{CJK*}(Meningotomy)" is 01.3100 and the code of "\begin{CJK*}{UTF8}{gbsn}丘脑切开术\end{CJK*}(Thalamus incision)" is 01.4101. The two terminologies belong to the same category because their code shares the same prefix "01". Therefore,for each mention, we first find the categories $C_m$ of is mapping terminologies, then randomly choose $k_n$ terminologies which belong to $C_m$.
    \item \textbf{Keywords replacing} We manually collect a keywords dictionary contains procedure site and procedure type. For each mention, we generate negative samples by replacing the keywords in its medical terminologies with other keywords in the dictionary. For instance, a medical mention "\begin{CJK*}{UTF8}{gbsn}甲状腺肿物摘除术\end{CJK*}(Meningotomy)" is mapped to terminology "\begin{CJK*}{UTF8}{gbsn}甲状腺病损切除术\end{CJK*}(Meningotomy)". The negative samples of keywords replacing could be "\begin{CJK*}{UTF8}{gbsn}肾上腺病损切除术\end{CJK*}(Meningotomy)" and "\begin{CJK*}{UTF8}{gbsn}甲状腺病损检查术\end{CJK*}(Meningotomy)". 
    \item \textbf{Online negative sampling} We choose the top-$k_n$ negative samples from the candidates of training set by the candidate generator after each training epoch. Random selection is used during the first epoch. The training process with online negative sampling is shown in Algorithm 1. 
\end{itemize}

The first three strategies choose the "hard negative samples" by heuristic rules. However, online negative sampling selects negative samples dynamically from the candidate generator. This end-to-end approach selects more suitable samples for the model to learn and improves its performance. The changing negative samples also provide diversity which enhanced the robustness for the model, which could be regarded an approach of data argumentation

\subsection{Keywords Attentive Ranker}
Keywords attentive ranker(KAR) aims to rank the candidates based on procedure sites and procedure types. The training data of KAR is generated by MTCG. We choose top-10 nearest terminologies for each mention as its candidates. KAR is trained in a point-wise way and the similarity is measured by the last output logits. The label is set to 1 if the mention is mapped to the terminology else 0.

To capture the keyword information, we propose a keyword attentive mechanism on Bert. Figure 5 shows the input of an example, we add two special token [PS] and [PT] at the beginning of the input sequence. In which PS stands for procedure site and PT stands for procedure type. Then the input of KAR is $s=\{[CLS],[PS],[PT],mention,[SEP],candidate,[SEP]\}$. Token [PS] only attends to procedure site words in mention and its candidate, token [PT] only pays attention to procedure type words in the sequence. This could be easily achieved by multiplying a self-attention mask in the transformer layer. More concretely, formula 3 could be written as formula 9, where $\cdot$ represent element-wise product. $Mask$ is the mask matrix and $Mask[i][j]=1$ if $s[i]$ attends to $s[j]$ else 0. The keywords are matched by a vocabulary collated by ourselves. [CLS] attends to the whole word in the input sequence and capture the semnatic information of mention and candidate in general, [PS] and [PT] only capture the procedure site and procedure type information respectively. We use an average pooling layer to merge the hidden output of [CLS], [PS] ,[PT] focus on different perspective. And followed by a FNN layer with a sigmoid function. The loss is calculated in formula 10-11, where $y$ is the label of mention and its candidates.

$$
head^i = (softmax(\frac{Q^iK^{iT}}{\sqrt{d_{head}}}) \cdot Mask)V^i\space\space\space\space\space\space\space\space\space\space\space\space\space\space(9)
$$
$$
loss = y\log{\hat{y}} + (1-y)\log{(1-\hat{y})}\space\space\space\space\space\space\space\space\space\space\space\space\space\space(10)
$$
$$
\hat{y} = KAR(mention, candidate)\space\space\space\space\space\space\space\space\space\space\space\space\space\space(11)
$$
\subsection{Fusion Block}
The fusion block is used in inference time, which merges the results of MTCG and KAR. First, given a medical mention $m$, MTCG recalls a candidates set $C$ contains $k_c$ terminologies. For each candidate $c_i$ in $C$, $d(m, c_i)$ stands for the euclidean distance of $m$ and $ith$ candidate in $C$, where $d(m, c_i) \in R$. Then KAR calculates the similarity with keywords information of $m$ and $c_i$, the output is notated as $sr_{m,c_i}$. The final similarity score $s_{m,c_i}$ of $m$ and $c_i$ takes both the similarity of the two models into consideration. $s_{m, c_i}$ is calculated as follows:
$$
s_{m,c_i} = \frac{sc_{m, c_i} +sr_{m,c_i}}{2}
$$

$$
sc_{m, C_i} = 1- \frac{d(m,c_i)}{\sum_{j=0}^{k_c}d(m, c_j)}
$$

Then the candidates $C$ is sorted by $s_{m,C_i}$ in descending order. Suppose the mention implication number predicted in MTCG is $x$ where $x \in [1,3]$. To handle the situation that the medical implication number is bigger than 3, we use a threshold $\theta$ to choose the possible terminologies. The final output terminologies is shown in formula 14, where $\theta$ is a hyper parameter.
$$
output = 
\begin{cases}
\{c_i|i\in[1, x]\} & {if\space\space x < 3} \\
\{c_i|s_{m,c_i}>\theta, i\in [4, |C|] \space or \space i \in [1, 3]\} & {if \space\space x=3}                
\end{cases} \space \space \space \space \space (14)
$$

\section{Experiments}
% In this section, to evaluate the performance of our proposed framework, we compare it with basic methods, search-rerank methods and generative-rerank methods. Besides, we also conduct extensive ablation study and case study to show the effectiveness of each block in the framework.

\subsection{Dataset}
The dataset is available at OpenKG, which is provided from the CHIP 2019 clinical entity normalization task. The statistic of training set and test set is shown in Table \ref{table1}. The mentions are collated from Chinese electronic health record and annotated by medical staff. And the knowledge base is ICD9-2017-PUMCH procedure codes(ICD9), which contains 9867 terminologies. The knowledge base is a tree structure and each terminology has its unique code. These terminologies are basically classified by procedure site which could be conducted by its code prefix.
\begin{table}
    \centering
    \caption{The statistics of train and test data}
    \label{table1}
    \begin{tabular}{@{}llll@{}}
    \hline
        Dataset &
        Uni-implication & Multi-implication & Total\\ \hline
        Train & 3801 & 199 & 4000\\
        Test & 2851 & 149 & 3000\\ \hline
    \end{tabular}
\end{table}

\subsection{Experiment settings}
We implement out model by pytorch and runs on 4 NVIDIA GeForce GTX 1080Ti GPUs. Owing to the high cost of pre-training Bert, we directly adopted BERT-base parameters pre-trained by Google in the Chinese general corpus. The optimizer of MTCG and KAR is both AdamW and they share the same learning rate 2e-5. For MTCG, the negative sample number for each mention $k_n$ is set to 4, the margin of triple loss is set to 1 and the candidate number $k_c$ generated during inference time is set to 10. In FB, the threshold $\theta$ is set to 0.8. Following the setting in \cite{yan2020knowledge}, we use 5-fold cross validation to measure the performance of out proposed model. We use accuracy to evaluate the performance of different methods. Considering the multi implication problem in this task, a sample is correct only when both the implication number and terminologies are exactly the same with the answer.

\subsection{Comparison with state-of-the-art methods}
We first compare our methods with two statistic methods: Tf-idf and edit-distance. A baseline Bert-rank model is also concluded. Besides, we also compare out methods with some state-of-the-art methods published in recent years. \cite{zhang2018effective} using combines different similarity algorithms based on n-gram as features to train a binary classifier to find similar symptoms. \cite{liang2020lab} is a "recall and rank" method which uses tf-idf to recall candidates and apply ESIM for similarity calculation in lab-indicator. However, both of their methods is not able to handle multi-implication thus we only choose the most similar one as the final output. \cite{yan2020knowledge} is a generative rerank method which propose a generative method for recalling and a Bert-base ranker for sort, and the implication number is generated during the first step. Note that the problem definition is not exactly same in different papers, so we slightly adjust these methods to satisfy our task for fair comparison.

The result is shown in Table \ref{table2}. It can be observed that our proposed method achieve the highest score in all metrics, with the uni-implication accuracy of 93.48$\%$, the multi-implication accuracy of 53.02$\%$ and the total acc of 91.47$\%$, which has a clearly improvements compared with other referenced methods. Tf-idf, edit-distance, \cite{zhang2018effective} and \cite{liang2020lab} performs poorly because they only consider the literature and semantic features, which is not enough for Chinese medical terminology normalization as analyzed in the first section. With the use of Bert, Bert-base rannking and \cite{yan2020knowledge} get high performance but still has a gap compared with our method. Besides, their methods are time consuming because bert-base ranking is a point-wise method and \cite{yan2020knowledge} generate candidates based on a sequence generative model.

\begin{table}
    \centering
    \caption{Performance of different methods in terms of accuracy}
    \label{table2}
    \begin{tabular}{@{}llll@{}}
    \hline
        Method & Uni-implication & Multi-implication & Total \\ \hline
        Tf-idf & 49.3 & - & 46.8 \\ 
        Edit-distance & 50.8 & - & 48.3 \\ 
        BERT-based ranking & 88.6 & - & 84.2 \\ 
        Zhang & 37.76 & - & 37.76 \\ 
        Liang & 69.63 & - & 69.63 \\ 
        Transformer & 91.1 & 52.4 & 89.3 \\ 
        Our methods & 93.48 & 53.02 & 91.47 \\ \hline
    \end{tabular}
\end{table}

\begin{figure}[h]
    \centerline{\includegraphics{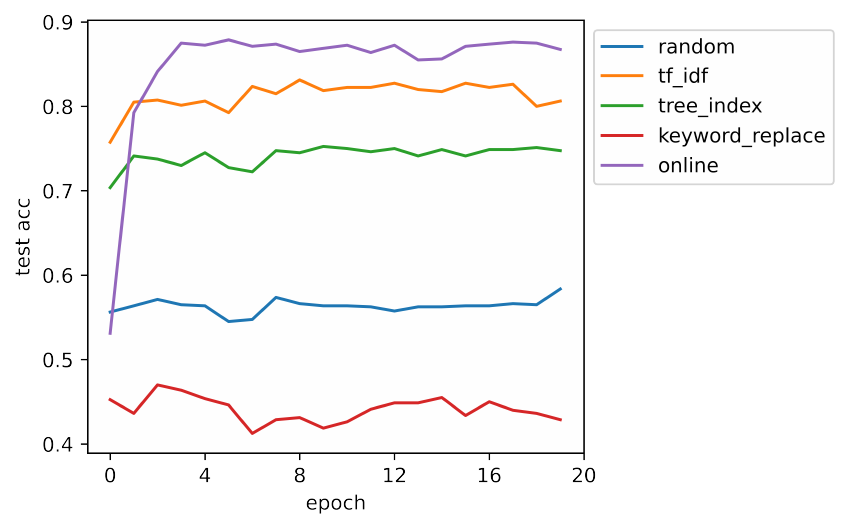}}
    \caption{Accuracy of different negative sampling strategies on test set during training process}
    \label{fig:4}
\end{figure}

\subsection{Ablation analysis}
To further investigate the importance of each block in our framework, we conduct an ablation study. As the shown in Table \ref{table3}, we have following observations: (1) MTCG plays an important role in our framework, without the high quality candidates generated by the recall model, the performance declines nearly 25$\%$, from 91.47$\%$ to 68.62$\%$. (2) KAR leads to a high performance on multi-implication. Because more than one procedure site or procedure type appears in multi-implication case. The candidate generator focuses more on literal and semantic similarity which ignores the keywords. Table \ref{table4} shows an example, where candidate generator ignores "\begin{CJK*}{UTF8}{gbsn}
肝\end{CJK*}" and the two generated terminologies are all relevant to "\begin{CJK*}{UTF8}{gbsn}胆囊\end{CJK*}". (3) Only use the output of MTCG or KAR could achieve relative high performance. However, their results focus on different perspective: literature similarity and keywords similarity, which is complementary for each other. As shown in Table \ref{table5}, the two cases is correct only when the result of MTCG and KAR all be taken into consideration. When the keywords in the medical mention is miss or unmatched with our vocabulary, the similarity comes from MTCG is a powerful complementary.

\begin{table}
    \centering
    \caption{Ablation study}
    \label{table3}
    \begin{tabular}{@{}llll@{}}
    \hline
        Method & Uni-implication & Multi-implication & Total \\ \hline
        Our methods & 93.48 & 53.02 & 91.47 \\ 
        w/o MTCG  & 72.21 & - & 68.62 \\ 
        w/o KAR & 92.63 & 41.61 & 90.1 \\ 
        w/o FB & 92.9 & 51.68 & 90.85 \\ \hline
    \end{tabular}
\end{table}

\begin{table*}
    \centering
    \caption{An normalization example. Our proposed KAR is able to focues on different procedure site}
    \label{table4}
    \begin{tabular}{@{}lll@{}}
    \hline
        Mention & Ca prediction & Kra prediction \\ \hline
        \begin{CJK*}{UTF8}{gbsn}肝癌切除术+胆囊切除术\end{CJK*} & \begin{CJK*}{UTF8}{gbsn}胆囊切除术\end{CJK*} & \begin{CJK*}{UTF8}{gbsn}肝病损切除术\end{CJK*} \\
         & \begin{CJK*}{UTF8}{gbsn}胆囊病损切除术\end{CJK*} & \begin{CJK*}{UTF8}{gbsn}胆囊切除术\end{CJK*} \\ \hline
    \end{tabular}
\end{table*}
\begin{table*}[h]
    \centering
    \caption{An normalization example. Our proposed method sort the terminologies from multi perspective thus lead to a high performance}
    \label{table5}
    \begin{tabular}{@{}lllll@{}}
    \hline
        Mention & Recall results & $sc_{m,c_i}$ & $sr_{m,c_i}$ & $s_{m,c_i}$ \\ \hline
        \begin{CJK*}{UTF8}{gbsn}腹腔镜胆囊切除术+术中胆道造影\end{CJK*} &
        \begin{CJK*}{UTF8}{gbsn}腹腔镜下胆囊切除术\end{CJK*} & 0.36 & 0.89 & 1.25 \\ 
         & \begin{CJK*}{UTF8}{gbsn}腹腔镜下胆囊造口术\end{CJK*} & 0.09 & 0.37 & 0.46 \\ 
         & \begin{CJK*}{UTF8}{gbsn}胆囊造影\end{CJK*} & 0.08 & 0.04 & 0.12 \\ 
         & \begin{CJK*}{UTF8}{gbsn}腹腔镜下胆道造影术\end{CJK*} & 0.06 & 0.61 & 0.67 \\ \hline
        \begin{CJK*}{UTF8}{gbsn}甲状舌骨囊肿切除术\end{CJK*} &  \begin{CJK*}{UTF8}{gbsn}甲状舌管病损切除术\end{CJK*} & 0.18 & 0.0001 & 0.1801 \\ 
         & \begin{CJK*}{UTF8}{gbsn}舌病损切除术\end{CJK*} & 0.17 & 0.0063 & 0.1763 \\ \hline
    \end{tabular}
\end{table*}

\subsection{Impacts of different negative sampling strategies}
In this section, we compare different negative sample strategies used in MTCG. As shown in Table \ref{table6}, online negative sampling has a huge improvement on accuracy compared with other strategies. Keywords replacing performs poor than random selection because it only replace the keywords, various expression examples are not presented during the training process. Tree coding and tf-idf find harder samples than random selection by statistic information, but their negative samples unchanged during each epoch. online negative sampling selects the most difficult samples in each epoch by the model as negative samples and achieves the best result. Figure 6 shows accuracy on test set at each epoch. Note that the accuracy of online negative sample is low because we use random sample in the first epoch. It is clear to see that with the use of online negative sampling, the accuracy has a rapid growth and leads other strategies with a large gap during the whole training process. 
\begin{table}
    \caption{Comparison of different negative sampling strategies}
    \label{table6}
    \centering
    \setlength{\tabcolsep}{7mm}{
    \begin{tabular}{@{}ll@{}}
    \hline
        Negative Sample Strategy & acc \\ \hline
        Keywords\_replace & 43.83 \\ 
        Random & 65.10 \\ 
        Tree-coding & 79.03 \\ 
        Tf-idf & 86.17 \\ 
        Online negative sampling & 90.10 \\ \hline
    \end{tabular}}
\end{table}
\subsection{Effectiveness and efficiency of recalling}
We explore the effectiveness and efficiency of recalling in this section. Here we use Recall@k as a metric to evaluate the recall performance. The result is shown in Table \ref{table7}. Compared with \cite{yan2020knowledge}, which utilize a generative model to recall. Our candidate generator based on deep metric learning performs better both in efficiency and effectiveness, especially an increase of nearly 42 times in speed. Because generative-based method is decoded word-by-word thus has a lower speed, our embedding-based model supports parallel computing and we use \cite{johnson2019billion} to quickly find the neighbors of a given embedding. Besides, we also experimented with different number of layers in Bert. As shown in the result, with the decrease of the number of layers, there is no obvious decrease in performance and a certain increase in speed, which prove that the deep metric learning is suitable for this task. The last two lines in Table \ref{table7} also shows online negative sampling is superior than tf-idf when the parameters decrease.

\begin{table*}[htbp]
    \centering
    \label{table7}
    \caption{The efficiency of different recall model}
    \begin{tabular}{@{}llllll@{}}
    \hline
        Methods & Recall@2 & Recall@5 & Recall@7 & Recall@10 & Speed \\ \hline
        Transformer & 88\% & 88.70\% & 89.10\% & 89.20\% & 12 mentions/sed \\ 
        Bert\_on\_12\_layers & 94.20\% & 97.23\% & 97.83\% & 98.30\% & 507 mentions/sed \\ 
        Bert\_on\_10\_layers & 93.63\% & 96.80\% & 97.40\% & 97.83\% & 588 mentions/sed \\ 
        Bert\_on\_8\_layers & 93.93\% & 96.93\% & 97.47\% & 97.97\% & 691 mentions/sed \\ 
        Bert\_on\_6\_layers & 93.23\% & 96.9\% & 97.43\% & 97.7\% & 833 mentions/sed \\ 
        Bert\_on\_2\_layers & 91.43\% & 95.07\% & 96.23\% & 96.57\% & 1500 mentions/sed \\ 
        Bert\_tf\_idf\_2\_layers & 83.93\% & 90.17\% & 91.33\% & 92.43\% & 1500 mentions/sed \\ \hline
    \end{tabular}
\end{table*}

\subsection{Implication number prediction}
In this section, we compare the accuracy of implication number prediction task. As shown in table \ref{table8}, the candidate generator performs best with the total accuracy of 98.23. What is interesting is that the model training separately on Bert performs poor than ours both in uni-implication and multi-implication, which indicates that the implication prediction task is affected by candidate generate: the representation contains enrich context used for recall is also benefit for implication number prediction.
\begin{table}[h]
    \centering
    \caption{The result of multi implication number prediction}
    \label{table8}
    \begin{tabular}{@{}llll@{}}
    \hline
        Method & Uni-implication & Multi-implication & Total \\ \hline
        BERT-based ranking & 100 & - & 96.3 \\ 
        delimiter + & 96.6 & 70.3 & 95.6 \\ 
        Transformer & 98.6 & 76.4 & 97 \\ 
        Bert-only & 99.26 & 65.77 & 97.8 \\ 
        Bert-multi-task & 99.58 & 70.71 & 98.23 \\ \hline
    \end{tabular}
\end{table}

\section{Discussion}
In this work, we first analysis four challenges for Chinese procedure normalization: multi-implication, short text, keywords sensitive and high efficiency. A novel framework is proposed which is composed of three parts. To handle multi implication problem and generate candidate terminologies quickly, we design a multi-task candidate generator(MTCG) based on deep metric learning. Online negative sampling strategy is used to generate hard samples for training which greatly improves recall rate. The keyword attentive ranker(KAR) is proposed to focus on procedure sites and procedure type of medical mentions and terminologies. A fusion block(FB) is also concluded to merge the results of MTCG and KAR. The experiments result shows our proposed framework outperforms existing methods and achieve the state of the art result on this task. In the future, we have great interests on the following three areas: 1) The candidate generator could be further speed up with tiny model architecture. 2) Automatically explore keywords without vocabulary. 3) Improve the performance with limited annotation data.

%%%%%%%%%%%%%%%%%%%%%%%%%%%%%%%%%%%%%%%%%%%%%%%%%%%%%%%%%%%%%%%%%%%%%%%%%%%%%%%%%%%%%
%
%     please remove the " % " symbol from \centerline{\includegraphics{fig01.eps}}
%     as it may ignore the figures.
%
%%%%%%%%%%%%%%%%%%%%%%%%%%%%%%%%%%%%%%%%%%%%%%%%%%%%%%%%%%%%%%%%%%%%%%%%%%%%%%%%%%%%%%

% \bibliographystyle{abbrv}
%\bibliographystyle{natbib}
% \bibliographystyle{achemnat}
% \bibliographystyle{plainnat}
% \bibliographystyle{bioinformatics}

%
\bibliographystyle{unsrt}
\bibliography{main}

% \begin{thebibliography}{}

% \end{thebibliography}

\end{document}